\documentclass[10pt,twocolumn,letterpaper]{article}

\usepackage{cvpr}              

\usepackage[dvipsnames]{xcolor}

\definecolor{cvprblue}{rgb}{0.21,0.49,0.74}
\usepackage[pagebackref,breaklinks,colorlinks,citecolor=cvprblue]{hyperref}

\def\paperID{13693} 
\def\confName{CVPR}
\def\confYear{2024}

\title{HOIST-Former: Hand-held Objects Identification, Segmentation, and Tracking in the Wild}

\author{
Supreeth Narasimhaswamy$^{1}$, Huy Anh Nguyen$^{1}$, Lihan Huang$^{1}$, and Minh Hoai$^{1,2}$ \\
$^{1}$Stony Brook University, USA, $^{2}$ VinAI Research, Vietnam
}

\begin{document}
\def\mA{\mathcal{A}}
\def\mB{\mathcal{B}}
\def\mC{\mathcal{C}}
\def\mD{\mathcal{D}}
\def\mE{\mathcal{E}}
\def\mF{\mathcal{F}}
\def\mG{\mathcal{G}}
\def\mH{\mathcal{H}}
\def\mI{\mathcal{I}}
\def\mJ{\mathcal{J}}
\def\mK{\mathcal{K}}
\def\mL{\mathcal{L}}
\def\mM{\mathcal{M}}
\def\mN{\mathcal{N}}
\def\mO{\mathcal{O}}
\def\mP{\mathcal{P}}
\def\mQ{\mathcal{Q}}
\def\mR{\mathcal{R}}
\def\mS{\mathcal{S}}
\def\mT{\mathcal{T}}
\def\mU{\mathcal{U}}
\def\mV{\mathcal{V}}
\def\mW{\mathcal{W}}
\def\mX{\mathcal{X}}
\def\mY{\mathcal{Y}}
\def\mZ{\mathcal{Z}} 

\def\bbN{\mathbb{N}} 
\def\bbR{\mathbb{R}} 
\def\bbP{\mathbb{P}} 
\def\bbQ{\mathbb{Q}} 
\def\bbE{\mathbb{E}}

\def\1n{\mathbf{1}_n}
\def\0{\mathbf{0}}
\def\1{\mathbf{1}}

\def\A{{\bf A}}
\def\B{{\bf B}}
\def\C{{\bf C}}
\def\D{{\bf D}}
\def\E{{\bf E}}
\def\F{{\bf F}}
\def\G{{\bf G}}
\def\H{{\bf H}}
\def\I{{\bf I}}
\def\J{{\bf J}}
\def\K{{\bf K}}
\def\L{{\bf L}}
\def\M{{\bf M}}
\def\N{{\bf N}}
\def\O{{\bf O}}
\def\P{{\bf P}}
\def\Q{{\bf Q}}
\def\R{{\bf R}}
\def\S{{\bf S}}
\def\T{{\bf T}}
\def\U{{\bf U}}
\def\V{{\bf V}}
\def\W{{\bf W}}
\def\X{{\bf X}}
\def\Y{{\bf Y}}
\def\Z{{\bf Z}}

\def\a{{\bf a}}
\def\b{{\bf b}}
\def\c{{\bf c}}
\def\d{{\bf d}}
\def\e{{\bf e}}
\def\f{{\bf f}}
\def\g{{\bf g}}
\def\h{{\bf h}}
\def\i{{\bf i}}
\def\j{{\bf j}}
\def\k{{\bf k}}
\def\l{{\bf l}}
\def\m{{\bf m}}
\def\n{{\bf n}}
\def\o{{\bf o}}
\def\p{{\bf p}}
\def\q{{\bf q}}
\def\r{{\bf r}}
\def\s{{\bf s}}
\def\t{{\bf t}}
\def\u{{\bf u}}
\def\v{{\bf v}}
\def\w{{\bf w}}
\def\x{{\bf x}}
\def\y{{\bf y}}
\def\z{{\bf z}}

\def\balpha{\mbox{\boldmath{$\alpha$}}}
\def\bbeta{\mbox{\boldmath{$\beta$}}}
\def\bdelta{\mbox{\boldmath{$\delta$}}}
\def\bgamma{\mbox{\boldmath{$\gamma$}}}
\def\blambda{\mbox{\boldmath{$\lambda$}}}
\def\bsigma{\mbox{\boldmath{$\sigma$}}}
\def\btheta{\mbox{\boldmath{$\theta$}}}
\def\bomega{\mbox{\boldmath{$\omega$}}}
\def\bxi{\mbox{\boldmath{$\xi$}}}
\def\bnu{\mbox{\boldmath{$\nu$}}}                                  
\def\bphi{\mbox{\boldmath{$\phi$}}}
\def\bmu{\mbox{\boldmath{$\mu$}}}

\def\bDelta{\mbox{\boldmath{$\Delta$}}}
\def\bOmega{\mbox{\boldmath{$\Omega$}}}
\def\bPhi{\mbox{\boldmath{$\Phi$}}}
\def\bLambda{\mbox{\boldmath{$\Lambda$}}}
\def\bSigma{\mbox{\boldmath{$\Sigma$}}}
\def\bGamma{\mbox{\boldmath{$\Gamma$}}}
                                  
\newcommand{\myprob}[1]{\mathop{\mathbb{P}}_{#1}}

\newcommand{\myexp}[1]{\mathop{\mathbb{E}}_{#1}}

\newcommand{\mydelta}[1]{1_{#1}}

\newcommand{\myminimum}[1]{\mathop{\textrm{minimum}}_{#1}}
\newcommand{\mymaximum}[1]{\mathop{\textrm{maximum}}_{#1}}    
\newcommand{\mymin}[1]{\mathop{\textrm{minimize}}_{#1}}
\newcommand{\mymax}[1]{\mathop{\textrm{maximize}}_{#1}}
\newcommand{\mymins}[1]{\mathop{\textrm{min.}}_{#1}}
\newcommand{\mymaxs}[1]{\mathop{\textrm{max.}}_{#1}}  
\newcommand{\myargmin}[1]{\mathop{\textrm{argmin}}_{#1}} 
\newcommand{\myargmax}[1]{\mathop{\textrm{argmax}}_{#1}} 
\newcommand{\myst}{\textrm{s.t. }}

\newcommand{\denselist}{\itemsep -1pt}
\newcommand{\sparselist}{\itemsep 1pt}

\definecolor{pink}{rgb}{0.9,0.5,0.5}
\definecolor{purple}{rgb}{0.5, 0.4, 0.8}   
\definecolor{gray}{rgb}{0.3, 0.3, 0.3}
\definecolor{mygreen}{rgb}{0.2, 0.6, 0.2}

\newcommand{\cyan}[1]{\textcolor{cyan}{#1}}
\newcommand{\blue}[1]{\textcolor{blue}{#1}}
\newcommand{\magenta}[1]{\textcolor{magenta}{#1}}
\newcommand{\pink}[1]{\textcolor{pink}{#1}}
\newcommand{\green}[1]{\textcolor{green}{#1}} 
\newcommand{\gray}[1]{\textcolor{gray}{#1}}    
\newcommand{\mygreen}[1]{\textcolor{mygreen}{#1}}    
\newcommand{\purple}[1]{\textcolor{purple}{#1}}       

\definecolor{greena}{rgb}{0.4, 0.5, 0.1}
\newcommand{\greena}[1]{\textcolor{greena}{#1}}

\definecolor{bluea}{rgb}{0, 0.4, 0.6}
\newcommand{\bluea}[1]{\textcolor{bluea}{#1}}
\definecolor{reda}{rgb}{0.6, 0.2, 0.1}
\newcommand{\reda}[1]{\textcolor{reda}{#1}}

\def\changemargin#1#2{\list{}{\rightmargin#2\leftmargin#1}\item[]}
\let\endchangemargin=\endlist
                                               
\newcommand{\cm}[1]{}

\newcommand{\mhoai}[1]{{\color{blue}{[MH: #1]}}}
\newcommand{\sn}[1]{{\color{red}{[SN: #1]}}}
\newcommand{\ha}[1]{{\color{orange}{[HA: #1]}}}

\newcommand{\mtodo}[1]{{\color{red}$\blacksquare$\textbf{[TODO: #1]}}}
\newcommand{\myheading}[1]{\vspace{1ex}\noindent \textbf{#1}}
\newcommand{\htimesw}[2]{\mbox{$#1$$\times$$#2$}}


\newif\ifshowsolution
\showsolutiontrue

\ifshowsolution  
\newcommand{\Comment}[1]{\paragraph{\bf $\bigstar $ COMMENT:} {\sf #1} \bigskip}
\newcommand{\Solution}[2]{\paragraph{\bf $\bigstar $ SOLUTION:} {\sf #2} }
\newcommand{\Mistake}[2]{\paragraph{\bf $\blacksquare$ COMMON MISTAKE #1:} {\sf #2} \bigskip}
\else
\newcommand{\Solution}[2]{\vspace{#1}}
\fi

\newcommand{\truefalse}{
\begin{enumerate}
	\item True
	\item False
\end{enumerate}
}

\newcommand{\yesno}{
\begin{enumerate}
	\item Yes
	\item No
\end{enumerate}
}

\newcommand{\Sref}[1]{Sec.~\ref{#1}}
\newcommand{\Eref}[1]{Eq.~(\ref{#1})}
\newcommand{\Fref}[1]{Fig.~\ref{#1}}
\newcommand{\Tref}[1]{Table~\ref{#1}}

\newcommand{\parens}[1]{\left(#1\right)}
\newcommand{\braces}[1]{\left\{#1\right\}}
\newcommand{\bracks}[1]{\left[#1\right]}
\newcommand{\modulus}[1]{\left\vert#1\right\vert}
\newcommand{\norm}[1]{\left\Vert#1\right\Vert}
\newcommand{\angular}[1]{\langle#1\rangle}
\newcommand{\lmod}{\left|\!\left|}
\newcommand{\rmod}{\right|\!\right|}

\twocolumn[{%
\renewcommand\twocolumn[1][]{#1}%
\maketitle
\begin{center}
    \centering
    \captionsetup{type=figure}
    \includegraphics[width=0.97\textwidth]{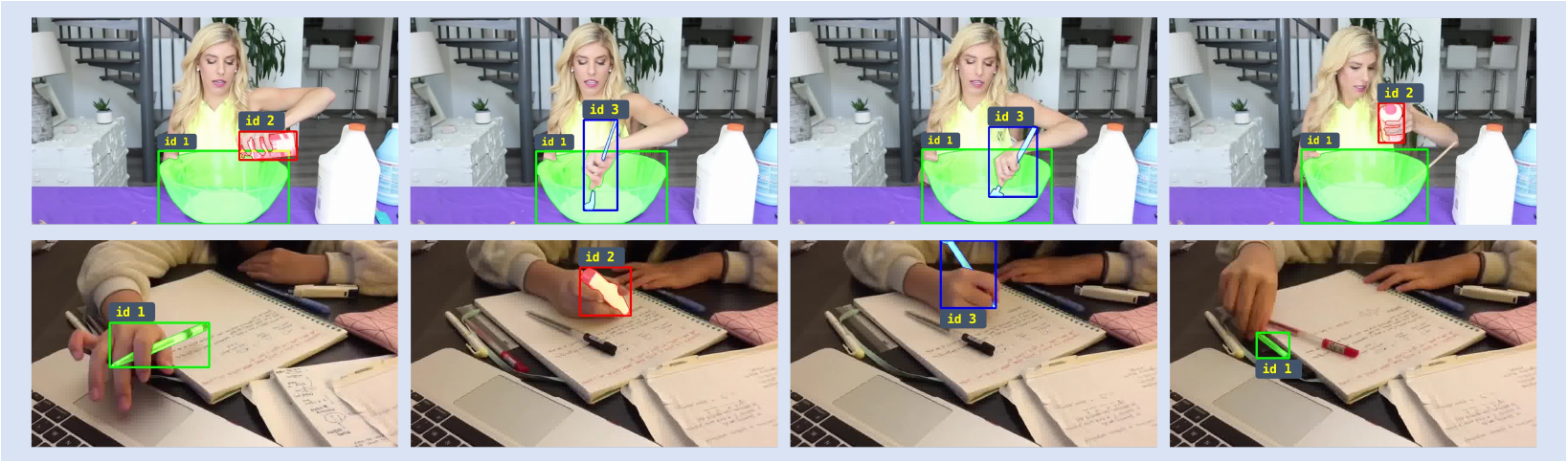}
    \vskip -.1in
    \caption{Identification, segmentation, and tracking of hand-held objects.}\label{fig.banner}
\end{center}%
}]

\begin{abstract}
\vskip -0.1in
We address the challenging task of identifying, segmenting, and tracking hand-held objects, which is crucial for applications such as human action segmentation and performance evaluation. This task is particularly challenging due to heavy occlusion, rapid motion, and the transitory nature of objects being hand-held, where an object may be held, released, and subsequently picked up again. To tackle these challenges, we have developed a novel transformer-based architecture called HOIST-Former. HOIST-Former is adept at spatially and temporally segmenting hands and objects by iteratively pooling features from each other, ensuring that the processes of identification,  segmentation, and tracking of hand-held objects depend on the hands' positions and their contextual appearance. We further refine HOIST-Former with a contact loss that focuses on areas where hands are in contact with objects. Moreover, we also contribute an in-the-wild video dataset called HOIST, which comprises 4,125 videos complete with bounding boxes, segmentation masks, and tracking IDs for hand-held objects. Through  experiments on the HOIST dataset and two additional public datasets, we demonstrate the efficacy of HOIST-Former in segmenting and tracking hand-held objects. Project page: \url{https://supreethn.github.io/research/hoistformer/index.html}

\end{abstract}

\section{Introduction}

Humans primarily use their hands to interact with their surroundings, making the ability to segment and track hand-held objects crucial for understanding and interpreting human interactions with the environment. From monitoring a factory worker navigating through assembly tasks to evaluating the skill set of a resident doctor performing intricate medical operations, the dynamic interplay between hands and objects forms the core of many activities. Segmenting hand-held objects allows computer vision systems to identify the focal points of action, while tracking these objects over time provides a coherent understanding of sequential and complex actions. This combined capability is particularly crucial in scenarios involving multiple similar objects, as it requires the system to differentiate and monitor the path of each item to deliver contextually rich, actionable insights.

In this paper, we study the problem of jointly segmenting and tracking objects that are held and moved by hands in unconstrained videos, as illustrated in \cref{fig.banner}. Specifically, given an input video composed of a sequence of frames, we consider all portable objects that are held by hands at any point within these frames. Suppose there are a number of such object instances. For each object instance, our objective is to produce a series of binary segmentation masks corresponding to each frame, such that the mask for a particular frame is empty if the object instance is not being held by a hand in that frame. Note that our study is limited to portable objects that can be held and moved by hand, excluding non-portable objects like furniture.

Segmenting and tracking hand-held objects involves three complex sub-tasks: first, identifying the object in the grasp of a hand from among several; second, accurately segmenting that object; and third, maintaining its track throughout the video.  Identifying hand-held objects is challenging because the mere overlapping of hand and object segments does not confirm a hold, as they may overlap in a 2D view without actual 3D contact. The segmentation task is complicated by heavy occlusion of objects by the hands, resulting in non-contiguous segments and the need to account for various object shapes and appearances in an open-world setting, regardless of category. Tracking is made difficult by the rapid movement of hands, which can drastically alter the position of the object from one frame to the next, potentially causing incorrect associations of object identity over time. Furthermore, while an object can be visible throughout the video, being hand-held is not a persistent characteristic; an object might be held at one moment, released the next, and picked up again subsequently. At times when the object is not in hand, segmentation and tracking should cease, notwithstanding its visibility. Despite these breaks in continuity, the object should maintain a consistent identifier throughout the video. Some of the challenges described here are illustrated in \cref{fig.banner}. 

To address the aforementioned challenges of \textbf{H}and-held \textbf{O}bjects \textbf{I}dentification \textbf{S}egmentation and \textbf{T}racking, we propose HOIST-Former. This model builds on the transformer-based image and video segmentation method Mask2Former ~\cite{cheng2021mask2former, Mask2Former_Video}, enhancing it with an innovative decoder architecture designed to overcome its limitations. Although Mask2Former is a leading method for object segmentation and tracking, its reliance on a predefined set of object categories makes it unsuitable for the segmentation and tracking of arbitrary hand-held objects in an open-world setting. Furthermore, Mask2Former's methodology, grounded solely on categorical membership and object visibility, is inadequate in scenarios where segmentation and tracking need to be initiated, paused, and resumed based on additional criteria, such as the hand-held status of an object, which is the central concern of this paper.

HOIST-Former addresses the limitations of Mask2Former with a novel Hand-Object Transformer decoder, which iteratively localizes hands and hand-held objects by mutually pooling features, effectively conditioning the identification and segmentation of the hand-held objects based on the appearance of hands and their surrounding context. Specifically, from a given set of video frames, a backbone network extracts low-resolution spatio-temporal features. These features are then gradually upscaled by a pixel decoder to produce high-resolution, per-pixel spatio-temporal embeddings. Finally, the Hand-Object Transformer decoder utilizes these high-resolution embeddings to generate spatio-temporal segmentation masks for both hands and the objects they are holding.

To train and evaluate HOIST-Former, we have collected and annotated a large-scale in-the-wild video dataset, named HOIST, a contribution of this work. Specifically, for each hand-held object in the video, we annotate its segmentation mask and assign a tracking instance ID that persists throughout the video. Our dataset comprises 4,228 videos with approximately 85,000 frames in total. The HOIST dataset includes numerous videos featuring hand-held objects within challenging and unconstrained environments, which can be used to train robust methods for hand-held object segmentation and tracking.

Experiments conducted on the HOIST dataset, along with two other datasets, reveal that HOIST-Former achieves superior results in the segmentation and tracking of hand-held objects.

\section{Related Work}

\myheading{Hand Analysis.} Hand analysis is crucial for many computer vision applications, so various problems have been studied. For example, there are works on detecting hands in images~\cite{wu_accv_2000, zhu_fg_2000, kolsch_fg_2004, ong_fg_2004, buehler_bmvc_2008, kumar_iccv_2009, mittal_bmvc_2011, pisharady_ijcv_2013, Narasimhaswamy_2019_ICCV}. There are also prior works that estimate hand contact~\cite{contacthands_2020, brahmbhatt_eccv_2020, mueller_cvpr_2021} and localize the contact objects in images using bounding boxes~\cite{shan_cvpr_2020}. Some works analyze hands by estimating their poses~\cite{zimmermann_iccv_2017, zimmermann_iccv_2019, hasson_cvpr_2019, romero_siggraph_2017, Liu_2021_CVPR, Chao_2021_CVPR, Rudnev_2021_ICCV, Kim_2021_ICCV, Yang_2021_ICCV, Kwon_2021_ICCV, Cao_2021_ICCV}, tracking them in videos~\cite{sridhar_cvpr_2015, zhang_arxiv_2020, wang_tog_2009, sharp_chi_2015, mueller_cvpr_2017, sridhar_eccv_2016, handler_2022}, and even generating them~\cite{sn_handdiffusion_iccv_2023, sn_handiffuser_cvpr_2024}. However, none of these works address segmenting and tracking hand-held objects.

Some works address the task of jointly estimating and tracking hand and object poses in videos~\cite{hampali2020honnotate, hampali2022keypointtransformer, H20_ICCV, DexYCB_CVPR}. However, they do not focus on segmenting or tracking hand-held objects as their main goal is pose estimation. Moreover, these methods typically deal with egocentric videos in constrained indoor environments with simple backgrounds and limited object categories. There is prior work on segmenting hands and hand-held objects~\cite{EgoHOS}, but tracking is not addressed. Our work is more related to video hand-held object segmentation \cite{VISOR_2022}, but this task requires segmentation masks of the objects in the first frame as an additional input and propagates these masks in subsequent frames. In contrast, our method only requires video frames as input to jointly segment and track hand-held objects. Another related problem is Video Instance Segmentation (VIS)~\cite{Yang2019vis}. However, VIS methods such as Mask2Former~\cite{Mask2Former_Video} segment and track objects from a predefined categories and are unsuitable for localizing arbitrary hand-held objects in open-world settings. Conversely, the proposed method HOIST-Former can segment and track arbitrary hand-held objects.

\myheading{Datasets.} 
One of our key contributions is the creation of a novel annotated dataset specifically designed for segmenting and tracking hand-held objects, which offers unique features not found in existing datasets. While many hand datasets are available, they are unsuitable for our purpose, either lacking video data~\cite{Narasimhaswamy_2019_ICCV, contacthands_2020, bodyhands_2022, shilkrot_bmvc_2019} or missing annotations for hand-held objects~\cite{handler_2022}. The 100DOH~\cite{shan_cvpr_2020} is a video dataset that includes bounding box annotations for hands and hand-contact objects, but it lacks tracking annotations and randomly samples frames for annotation. EgoHOS~\cite{EgoHOS} provides segmentation masks for hands and hand-held objects in frames from egocentric videos. However, it does not offer consistent tracking annotations, assigning the same instance ID to different objects held by a hand at different times, thereby limiting its utility for our purpose. 

Our dataset shares similarities with the VISOR benchmark~\cite{VISOR_2022}, which annotates Epic-Kitchen videos~\cite{EpicKitchens_2021} with segmentation masks and tracking IDs for objects interacting with hands. However, VISOR includes all types of interacting objects, not just those held by hands. Another limitation of VISOR is its focus solely on egocentric kitchen videos, whereas our HOIST dataset includes a wide range of diverse, unconstrained videos from various settings. While there are general-purpose video object segmentation datasets like~\cite{Yang2019vis, xu2018youtube, UVO}, they do not specifically annotate hand-held objects; we manually selected 80 videos featuring hand-held objects from these datasets for additional evaluation. However, this quantity is insufficient for training, highlighting the necessity of our dataset.

\section{HOIST-Former}

This section describes HOIST-Former, a novel network designed to jointly segment and track hand-held objects. The network takes as input a video $\mathcal{V} \in \bbR^{T{\times}H{\times}W{\times}3}$, consisting of $T$ frames with spatial dimensions $H{\times}W$ and three color channels. For each object $O$ in the video that is held by a hand in at least one of the frames, the network outputs a binary 3D tensor $\M \in \{0,1\}^{T{\times}H{\times}W}$, representing the spatio-temporal locations of the object. If object $O$ is not held by a hand in a frame, then its corresponding 2D binary mask will be devoid of any segment, as our task is solely on segmenting and tracking the object when it is being held. Note that object $O$ retains a unique instance ID throughout the video, even if certain 2D segmentation masks become empty at times due to complete occlusion or temporary disruption of the hand-held status. 

\begin{figure}
\centering 
    \centering
    \includegraphics[width=0.99\linewidth]{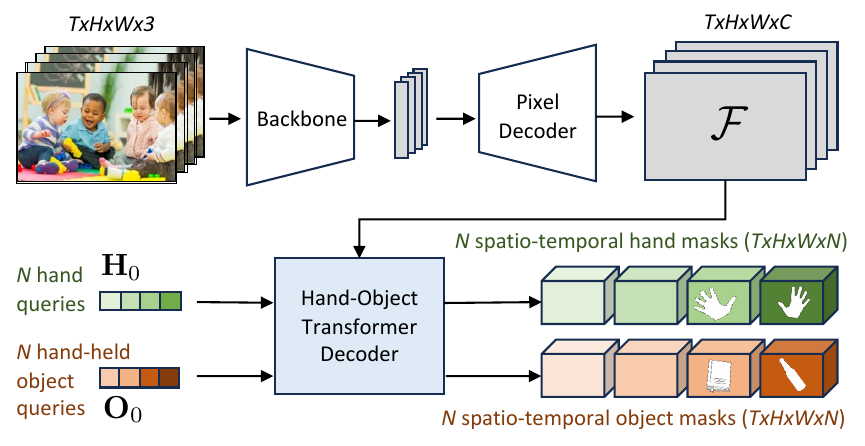}
    \vskip -0.1in
    \caption{HOIST-Former consists of a backbone network, a pixel decoder, and a transformer decoder. The input video is initially processed through the backbone network and the pixel decoder to generate high-resolution spatio-temporal features $\mF$. The transformer decoder operates on $\mF$, decoding a set of N hand queries and their corresponding object queries, resulting in N spatio-temporal hand masks and corresponding object masks.} \label{fig.architecture}
    \vskip -0.1in

\end{figure}

\begin{figure*}
\centering 
    \centering
    \includegraphics[width=0.95\textwidth]{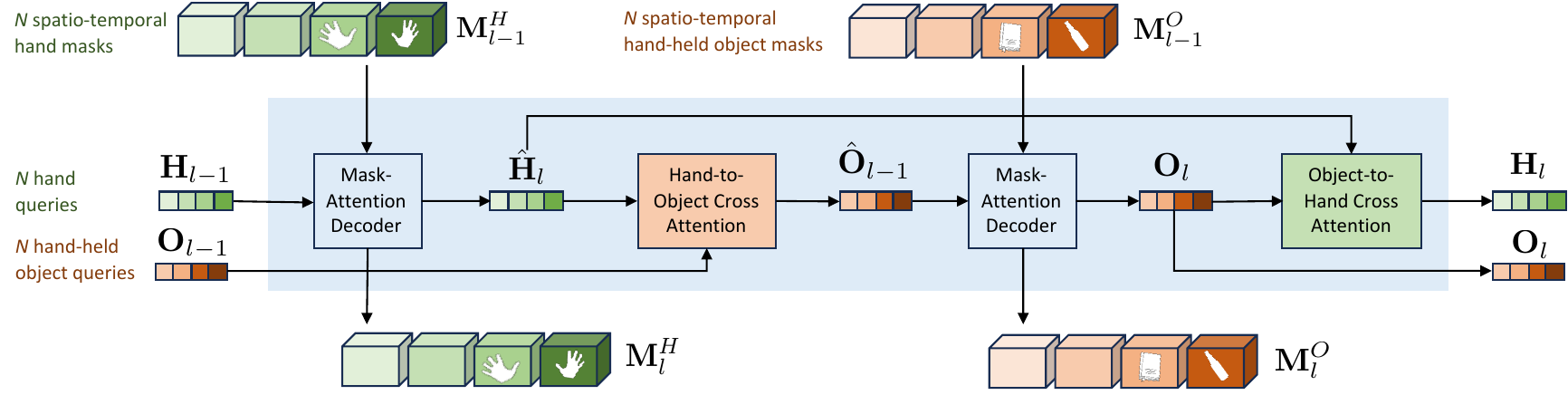}
    \caption{The Hand-Object Transformer Decoder features a network architecture with $L$ layers. This figure demonstrates the operational flow of a single layer, which includes two mask attention modules and two cross-attention modules. The inputs for this layer consist of N sets of four elements each: a hand query, an object query, a spatio-temporal hand mask, and a spatio-temporal object mask. The outputs of this layer are the correspondingly updated versions of these entities.} \label{fig.decoder}
    \vskip -0.1in
\end{figure*}

Inspired by the success of Mask2Former in video instance segmentation, we designed HOIST-Former with a similar overall architectural framework, consisting of three main components: a backbone network, a pixel decoder, and a transformer decoder, as depicted in \Fref{fig.architecture}. First, the input video is fed into the backbone network to extract low-resolution features. These features are then upscaled by the pixel decoder to generate high-resolution spatio-temporal features $\mF$. The transformer decoder operates on $\mF$, processing hand and object queries iteratively. These queries, starting as an initial set of learnable $C$-dimensional feature vectors representing potential hands or hand-held objects in the video, are iteratively updated by the transformer decoder. The spatio-temporal binary mask predictions for both object and hand tracks are decoded from these hand and object queries in conjunction with the high-resolution spatio-temporal features~$\mF$.

In the remainder of this section, we will describe the innovative transformer decoder of  HOIST-Former, called Hand-Object Transformer Decoder. Following this, we will describe how HOIST-Former can be trained. 

\subsection{Hand-Object Transformer Decoder} 

The Hand-Object Transformer Decoder is an innovative component, designed to systematically determine the positions of hands and hand-held objects through an iterative and collaborative feature pooling process. This effectively conditions the identification and segmentation of hand-held objects based on the appearance of hands and their immediate environment. This innovative transformer decoder allows us to segment and track arbitrary hand-held objects in an open-world setting, satisfying selection criteria that extend beyond categorical membership and object visibility.

Given the spatio-temporal features $\mathcal{F}$, we start with $N$ learnable hand queries $\H_0 \in \mathbb{R}^{N{\times} C}$ and object queries $\O_0 \in \mathbb{R}^{N{\times} C}$. These queries function similarly to region proposals~\cite{ren_nips_2015} and can generate spatio-temporal segmentation masks for hands and hand-held objects by attending to the features $\mathcal{F}$. Similar to Mask2Former~\cite{Mask2Former_Video}, these queries are processed by $L$ transformer decoder layers to produce segmentation masks. However, unlike Mask2Former, our focus is on segmenting objects based on their interaction with hands, regardless of their category or visibility. We therefore condition object segmentation on the appearance of hands and their surrounding context. Conversely, identifying hand-held objects aids in localizing hands. Therefore, we condition the hand segmentation upon hand-held objects. These dual tasks are achieved by mutually pooling information between hand and object queries.

The operational flow of the Hand-Object Transformer Decoder is illustrated in \Fref{fig.decoder}. It encompasses four principal operations: an initial mask attention operation, succeeded by a cross-attention operation, another mask attention operation, and finally concluding with a second cross-attention operation. The formal representation of these four steps is provided in the following equations:
\begin{align}
&\hat{\H}_l, \M^{H}_{l} := MaskAtt(\H_{l-1}, \M^{H}_{l-1}\  | \  f^{H \rightarrow H}), \\
&\hat{\O}_{l-1} := CrossAtt(\O_{l-1}, \hat{\H}_l \  | \  f^{H \rightarrow O}),  \\
&\O_l, \M^{O}_{l} := MaskAtt(\hat{\O}_{l-1}, \M^{O}_{l-1}\ | \  f^{O \rightarrow O}),  \\
&\H_{l} := CrossAtt(\hat{\H}_{l}, \O_{l}\ |\  f^{O \rightarrow H}) 
\end{align}
In the above, $CrossAtt$ and $MaskAtt$ refer to the cross-attention and mask-attention modules, respectively. To elaborate, the function $CrossAtt(\X, \Y | f)$ takes two inputs, $X$ and $Y$, where $f$ symbolizes a trio of linear functions  $f_Q(\cdot), f_K(\cdot), f_V(\cdot)$ (for query, key, value) with learnable parameters. The function $CrossAtt$ draws information from $Y$ to $X$, resulting in an updated version $\X'$ of $\X$. This process is defined as follows:
\begin{align}
    \X' = softmax\parens{f_Q(\X)f_K(\Y)^T}f_V(\Y). 
\end{align}

The function $MaskAtt(\X, \M | f)$ operates with two inputs: the query set $X$ and the collection of spatio-temporal binary masks $\M$. It outputs the revised queries $X'$ and the updated masks $\M'$. Also in this context, $f$ represents a set of three linear functions: $f_Q(\cdot), f_K(\cdot)$, and $f_V(\cdot)$, each characterized by learnable parameters. The update equation for the queries is: 
\begin{align}
\X' = softmax(\mM + f_Q(\X)f_K(\mF)^T)f_V(\mF) + \X. 
\end{align}
In the above, $\mF$ is the high-resolution spatio-temporal feature maps from the pixel decoder. The 4D attention mask $\mM$ is determined by the set of 3D binary masks $\M$. The value of $\mM$ at location $(t, y, x, n)$ is: 
\begin{align}
    \mM(t, y, x, n) = \begin{cases}
        0 & \text{if } \M(t, y, x, n) = 1,\\
        - \infty & \text{otherwise}.
    \end{cases} \label{eqn:msk_hand}
\end{align}

Another output of the $MaskAtt(\X, \M | f)$ function is the set of updated 3D masks $\M'$, which are obtained by using dot products between query features $\X$ and spatio-temporal features $\mF$. We refer the reader to~\citet{Mask2Former_Video} for more details.

Note that the Hand-Object Transformer Decoder is composed of four Attention modules: two cross-attention and two mask-attention modules. Each of these modules is equipped with three linear functions corresponding to query, key, and value, resulting in a total of 12 linear functions, each featuring learnable weights.

\begin{figure*}
\centering 
    \centering
    \includegraphics[width=0.23\linewidth]{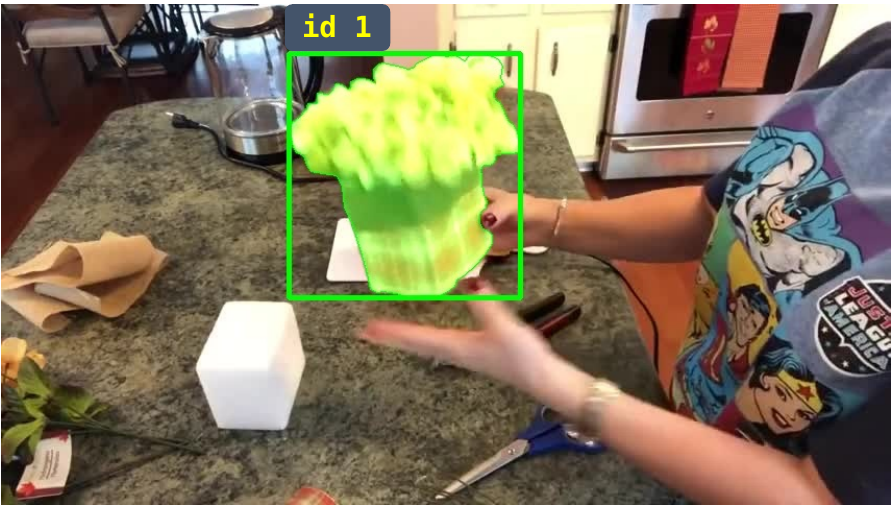}
    \includegraphics[width=0.23\linewidth]{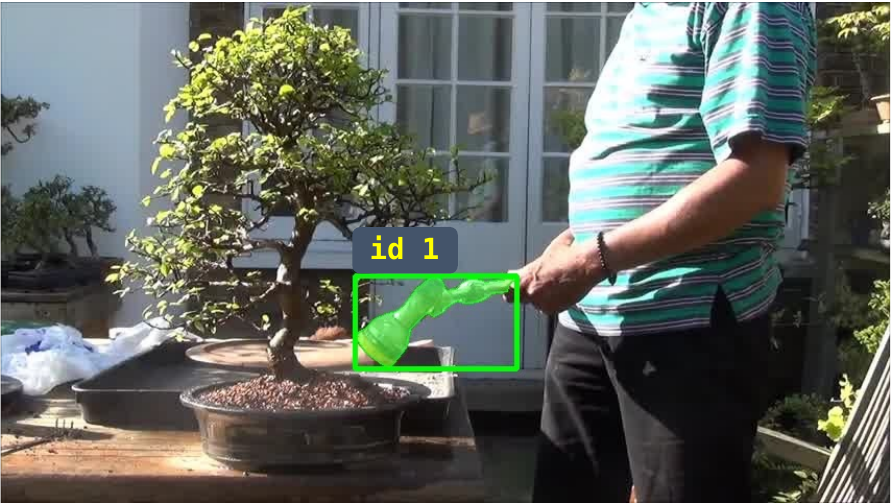}
    \includegraphics[width=0.23\linewidth]{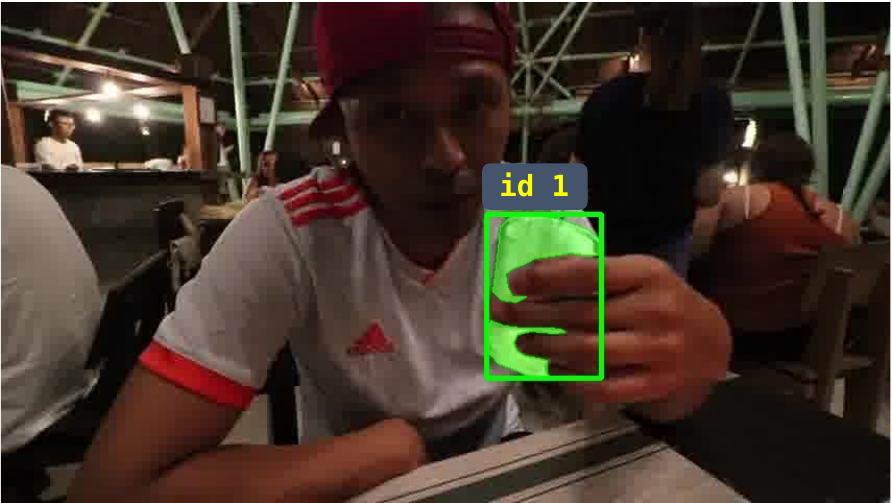}
    \includegraphics[width=0.23\linewidth]{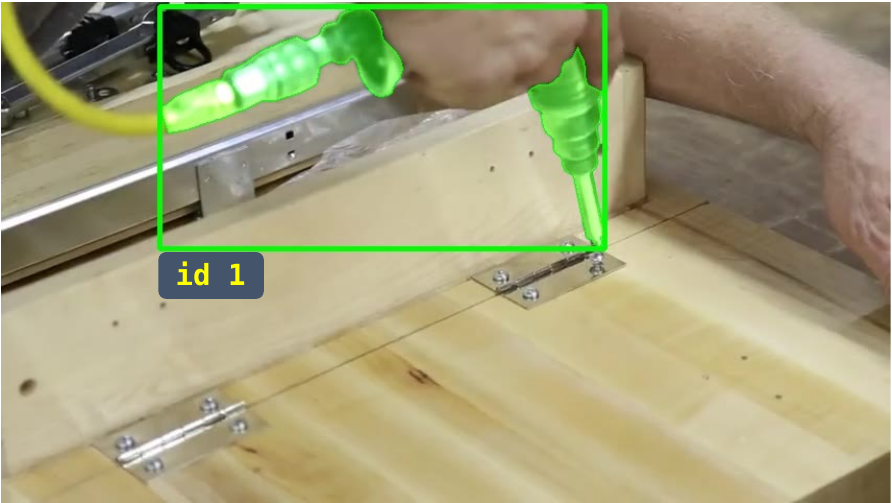}
    \includegraphics[width=0.23\linewidth]{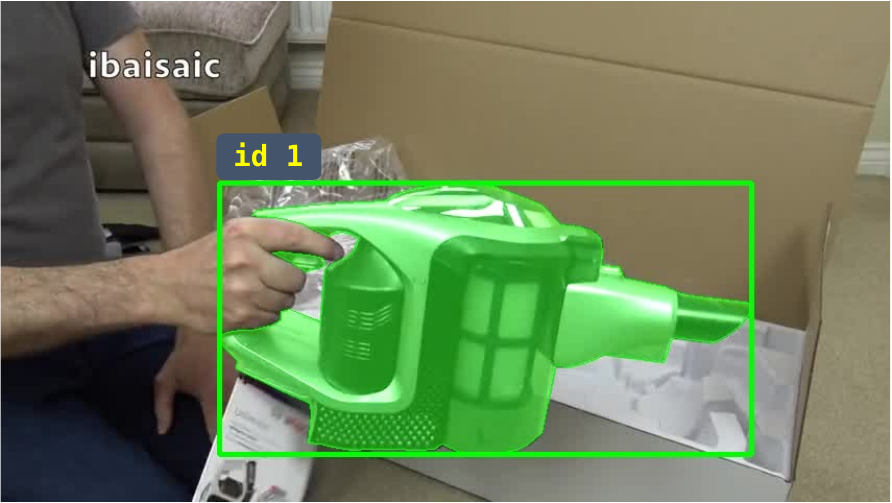}
    \includegraphics[width=0.23\linewidth]{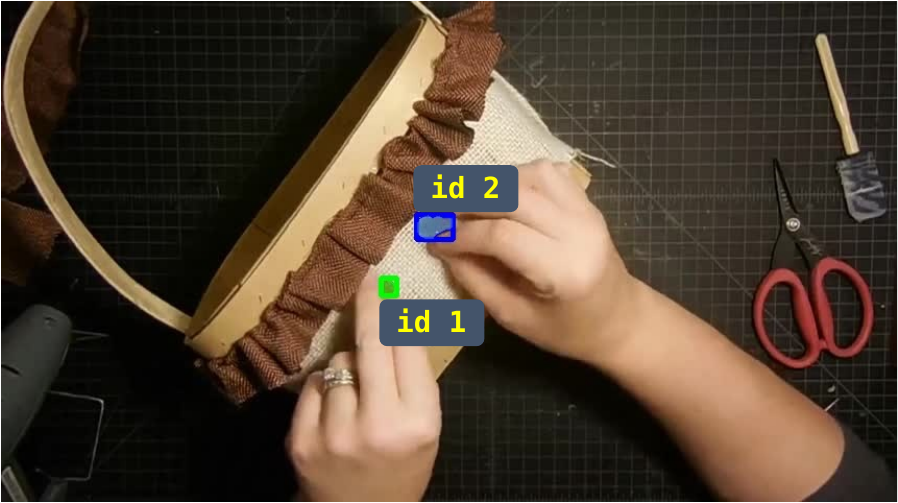}
    \includegraphics[width=0.23\linewidth]{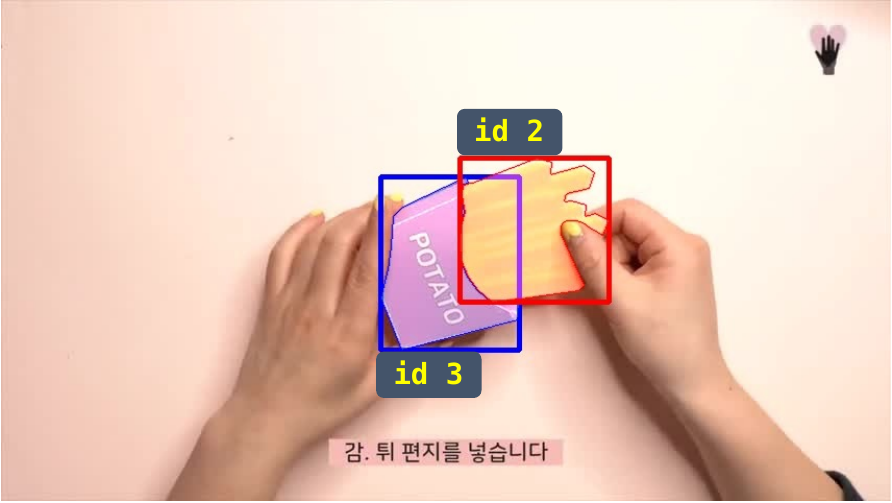}
    \includegraphics[width=0.23\linewidth]{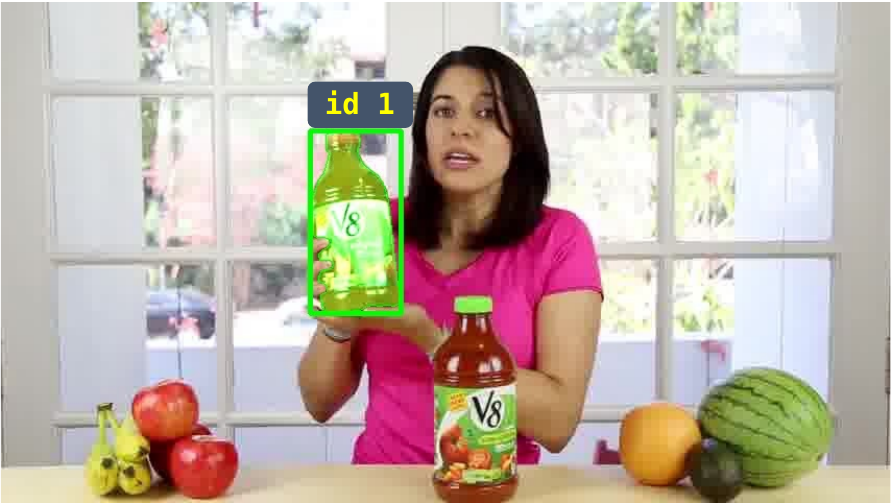}
    \caption{{\bf Sample frames from HOIST.} HOIST dataset contains videos with diverse scenes, camera views, object sizes, and occlusions.} \label{fig.HOIST_dataset}
    \vskip -0.1in

\end{figure*}

\subsection{Training Losses}

We train HOIST-Former using the  multi-task loss:

\begin{equation}
    \mL = \mL_{cls} + \mL_{mask} + \mL_{dice}.\label{eqn:loss}
\end{equation}

The term $\mL_{cls}$, $\mL_{mask}$, and $\mL_{dice}$  denote the class loss, mask loss, and dice losses, respectively. Both the mask loss and dice loss comprise a linear combination of individual losses calculated for hands, objects, and contact masks
\begin{align}
    & \mL_{mask} = \lambda_1 \mL^{H}_{mask} + \lambda_2 \mL^{O}_{mask} + \lambda_3 \mL^{C}_{mask}, \label{eqn:loss_mask} \\
    &\mL_{dice} = \lambda_4 \mL^{H}_{dice} + \lambda_5 \mL^{O}_{dice} + \lambda_6 \mL^{C}_{dice}.\label{eqn:loss_dice}
\end{align}

The mask loss and dice loss each include a loss term specific to the contact mask, which represents the interaction area between a hand-held object and the hand holding it. This contact mask is derived after acquiring the spatio-temporal hand and object masks at each decoder layer $l$. Encoding the key interaction zone between hands and objects, the contact mask plays a vital role in accurately localizing hand-held objects. During training, we incorporate contact losses to guide HOIST-Former's focus towards areas where hands and objects make contact. The effectiveness of including contact losses is demonstrated through empirical results in our experimental section.

The $\lambda_i$'s in \Eref{eqn:loss_mask} and \Eref{eqn:loss_dice} are tunable hyperparameters that control the relative strength of individual losses.

\section{HOIST Dataset}

This section describes a novel and challenging dataset we have collected to develop and evaluate hand-held object segmentation and tracking methods.

\myheading{Dataset Source.} 
Our goal is to compile a sufficiently large and diverse video dataset to develop methods for detecting, segmenting, and tracking hand-held objects. Specifically, we aim for the dataset to satisfy several criteria. First, it should include videos of everyday activities in diverse indoor and outdoor environments. Second, the videos should feature people interacting with a wide range of objects. Third, we aim to develop methods suitable for unconstrained videos showcasing hands interacting with multiple objects, involving scenarios where objects are held, released, and picked up in various sequences; hence, the dataset should include such types of videos. To fulfill these requirements, we selected YouTube videos from the 100DOH dataset~\cite{shan_cvpr_2020}, which consists of 100K frames from 27.3K videos. While 100DOH provides bounding box annotations for hands and hand-contact objects, it lacks segmentation or tracking annotations.

To construct HOIST, we initially focused on 100DOH frames featuring portable object annotations. We identified the publicly available YouTube videos corresponding to these frames as potential candidates. For each selected frame, we used shot boundary information to locate contiguous segments within the corresponding video, eliminating any duplicate segments in the process. The videos were post-processed to an approximate length of three seconds each. We extracted the videos at a rate of six frames per second, resulting in about 18 frames per video, and resized the frames to maintain a shorter side of 480 pixels. In total, our dataset comprises 4,228 videos with 83,970 frames.

\myheading{Annotation and Statistics.} In the HOIST dataset, we annotate every instance of hand-held objects in the videos. Specifically, we provide annotations for each object's bounding box, segmentation mask, and a unique instance ID to facilitate tracking. We omit the object's bounding box and mask annotations in frames where the object is not held by a hand. The annotation process begins with the manual annotation of bounding boxes and tracking IDs for hand-held objects. Subsequently, we divide the videos into train, validation, and test sets, adhering to the splits defined in the 100DOH dataset. For the test and validation videos, we manually annotate the segmentation masks of hand-held objects. In contrast, for the training videos, we generate segmentation masks by applying the Segment Anything~\cite{SAM_ICCV} model to the manually annotated bounding boxes. \Tref{table.data_statistics} provides some key statistics of the HOIST dataset and \Fref{fig.HOIST_dataset} illustrates some sample annotated frames.

\setlength{\tabcolsep}{3pt}
\begin{table}
\centering

\begin{tabular}{lrrrr}
\toprule
 & Train & Valid & Test & Total \\ 
\midrule 

\# Videos with no object  & 345    & 0     & 0 & 345   \\
\# Videos with 1 object   & 1896   & 99   & 178 & 2173 \\
\# Videos with 2 objects  & 1117   & 61    & 100 & 1278  \\
\# Videos with $\geq 3$ objects  & 367    &  22    & 43 & 432  \\
\# Videos - total                & 3725   & 182   & 321 & 4228 \\
\# Frames                 & 74527  & 3470  & 5973 & 83970 \\
\# Object Instances       & 5393   & 310   & 522 & 6225 \\

\bottomrule
\end{tabular}
\vskip -0.1in
\caption{\textbf{HOIST dataset statistics.}}\label{table.data_statistics}
\vskip -0.1in

\end{table}

\section{Experiments}


\setlength{\tabcolsep}{8pt}
\begin{table*}[ht]

\begin{center}
\begin{tabular}{lllllrrr}
\toprule
 &\multicolumn{2}{c}{First identification} &\multicolumn{2}{c}{Continued tracking at Frame $t$} & \multicolumn{3}{c}{Evaluation dataset}\\ 
\cmidrule(lr){2-3} \cmidrule(lr){4-5} \cmidrule(lr){6-8}
Method & Box  & Mask  & Box  & Mask & HOIST & VISOR & UVO \\ 
\midrule 
BL-A & N/A & GT &  N/A & STCN                                          & 48.9      & 33.1      & 40.4 \\
BL-B & GT & SAM &  N/A & STCN                                      & 41.6      & 23.7      & 33.7 \\
BL-C & 100DOH & SAM &  N/A & STCN                            & 23.3      & 8.0       & 15.9 \\
BL-D & GT & SAM & GT + IoUTracker & SAM                                            & 4.2       & 10.0      & 1.8  \\
BL-E & 100DOH & SAM & 100DOH + IoUTracker & SAM    & 1.2       & 5.2       & 1.3  \\
BL-F & GT & SAM & GT + StrongSORT & SAM                                            & 20.2      & 1.3       & 14.5 \\
BL-G & 100DOH & SAM &  100DOH + StrongSORT & SAM                & 4.7       & 0.5       & 2.0 \\
BL-H & GT & SAM & GT + StrongSORT++ & SAM                                          & 19.3      & 1.3       & 14.4 \\
BL-I & 100DOH & SAM &  100DOH + StrongSORT++ & SAM                                 & 5.4       & 0.5       & 3.5 \\
BL-J & GT & SAM & GT + ByteTrack & SAM                                            & 26.0      & 8.1       & 18.4 \\
BL-K & 100DOH & SAM &  100DOH + ByteTrack & SAM                & 7.8       & 3.2       & 2.8 \\
BL-L & GT & SAM & GT + MixFormer & SAM                           & 39.7      & 24.0      & 29.7 \\
BL-M & 100DOH & SAM &  100DOH + MixFormer & SAM                                    & 22.0      & 11.0      & 18.1 \\
BL-N & GT & SAM & GT + GTR & SAM                           & 25.8      & 21.3      & 21.8 \\

Mask2Former & N/A & N/A & N/A & N/A & 51.8 & 42.8 & 63.1 \\
HOIST-Former & N/A & N/A & N/A & N/A & {\bf 56.4} & {\bf 46.6} & {\bf 66.7} \\
\bottomrule
\end{tabular}
\end{center}
\vskip -0.15in 
\caption{Average Precision (AP) for diverse approaches to identifying, segmenting, and tracking hand-held objects. The first 14 methods (from BL-A to BL-N) vary in their strategies for initially identifying and then continuously tracking the bounding box and segmentation mask at Frame $t$. We consider several alternatives, from automatic methods that leverage state-of-the-art techniques in hand-held object detection (100DOH \cite{shan_cvpr_2020}), object segmentation (SAM~\cite{SAM_ICCV}), segmentation mask propagation (STCN \cite{stcn}), and bounding box tracking (IoUTracker \cite{ioutracker,he2020fastreid}, StrongSORT \cite{strongsort}, ByteTrack \cite{zhang2022bytetrack}, MixFormer \cite{mixformer}, 
GTR \cite{gtr}), to hypothetical scenarios where an oracle supplies ground truth bounding boxes or segmentation masks. This comprehensive analysis results in an extensive list of methods, as detailed in the table; however, all are outperformed by HOIST-Former by a wide margin.
\label{table.baselines}
}
\vskip -0.1in

\end{table*}

\setlength{\tabcolsep}{8pt}
\begin{table}
\begin{center}
\scalebox{1.0}{
\begin{tabular}{lc}
\toprule
Method & AP \\
\midrule
HOIST-Former & 56.4 \\
HOIST-Former w/o Hand-to-Object Attn. & 52.7 \\
HOIST-Former w/o Object-to-Hand Attn. & 53.5 \\
HOIST-Former w/o Contact Loss & 54.8 \\
\bottomrule
\end{tabular}}

\end{center}
\vskip -0.2in 
\caption{\textbf{Ablation study outcomes} for HOIST-Former evaluated on the HOIST dataset's test data. Removing either the Hand-to-Object Attention module, 
 the Object-to-Hand Attention module, or the Contact Loss would significantly degrade the performance of HOIST-Former. \label{table.ablation_studies}}
\vskip -0.2in

\end{table}

This section outlines the extensive experiments we conducted on several datasets. We begin by describing the datasets used, the evaluation metrics employed, and the training details.

\subsection{Datasets}

In addition to evaluating our method's effectiveness on the proposed HOIST dataset, we also perform experiments on selected videos from the VISOR~\cite{VISOR_2022} and UVO~\cite{UVO} datasets that are amendable for hands and hand-held objects evaluation. This section details the filtering steps we employed to select and prepare these videos.

\myheading{VISOR} is an egocentric video semantic segmentation dataset derived from EPIC-KITCHEN \cite{EpicKitchens_2021}, centered around active objects involved in the user's  actions. The dataset offers two types of annotations: manually curated, high-quality sparse annotations, and dense annotations generated through interpolation. Due to the unreliability of dense annotations, we only use the sparse annotations.

In its sparse annotations, each VISOR video clip typically encompasses three actions with a total of six annotated frames. From the annotations, we identified objects in direct contact with hands. Initially, we manually reviewed the object categories and names, excluding immobile categories like stovetops, dishwashers, and freezers. For certain remaining categories, we applied a further filter to eliminate excessively large objects, setting the mask area threshold for an ``overly large'' object at 0.3. Upon a detailed review of the dataset, we corrected several annotation errors, particularly those involving gloves and hands. The final step is to exclude object instances not in contact with hands by checking the overlap between diluted hands and instance masks. The VISOR dataset is divided into two subsets: originally, the Train subset had 5322 clips, and the Valid subset had 1251 clips. After our filtering steps, we are left with 5022 clips in the Train subset, featuring 31.4K frames and 34.5K hand-held object instances with mask annotations, and 1162 clips in the Valid subset, encompassing 7.3K frames with 8K hand-held object instances.

\myheading{UVO} is a general-purpose video segmentation dataset featuring annotations for various object categories, along with their segmentation masks and tracking IDs. To extract videos containing hand-held objects for segmentation and tracking, we employ a three-step process. First, we select a subset of UVO object categories likely to be hand-held, excluding categories like vehicles, people, and furniture. Second, we manually review videos under these potential hand-held object categories, retaining only those that actually feature hand-held objects. Third, we adjust the segmentation masks of these hand-held objects, setting them to empty masks in frames where they are not held by hands. This process results in 90 videos from the training set and 80 from the validation set. Given the insufficiency of 90 videos for training purposes, we opt to use only the 80 videos from the validation set for evaluation.

\subsection{Evaluation metric and training details}

We measure the joint performance of hand-held object segmentation and tracking using the Average Precision (AP). We consider a detected spatio-temporal mask $\M$ as a true positive if its Intersection over Union (IoU) with a ground-truth spatio-temporal mask is greater than 0.5. Specifically, given a detected spatio-temporal mask $\M=(\M_1, \M_2,\cdots, \M_T)$ and a ground-truth spatio-temporal mask $\M^{gt}=(\M^{gt}_1, \M^{gt}_2,\cdots, \M^{gt}_T)$, we compute the IoU as follows:
\begin{equation}
    \text{IoU}\parens{\M, \M^{gt}} = \dfrac{\sum_{t=1}^T \left|\M_t \cap \M^{gt}_{t} \right|}{\sum_{t=1}^T \left|\M_t \cup \M^{gt}_{t} \right|}.
\end{equation}

We implement HOIST-Former using Detectron2~\cite{wu2019detectron2}. We set the hyperparmeters  $\lambda_2, \lambda_5$ in \Eref{eqn:loss_mask} and \Eref{eqn:loss_dice} to be 5. We set the rest of $\lambda_i$'s to be 0.001. We train HOIST-Former on eight 80GB GPUs using AdamW optimizer with an initial learning rate of $0.0001$.

\subsection{Comparison Methods \label{sec:compare-methods}}

Our objective is to effectively identify, segment, and track hand-held objects. Any viable approach must address these functions in some manner. We evaluate HOIST-Former against a comprehensive range of methods, each representing different approaches to these tasks.

The first task involves locating a hand-held object's bounding box. The second requires generating a segmentation mask, and the third entails maintaining identity consistency across frames, either by matching detected object instances or through propagation. These tasks hinge on the initial identification and subsequent tracking of the bounding box and segmentation mask at Frame $t$. We consider two options for obtaining the initial bounding box: automatic detection using the 100DOH detector \cite{shan_cvpr_2020}, or using a human-annotated ground truth bounding box. For initial segmentation, one option is using SAM with the bounding box as a prompt; alternatively, methods that use ground truth masks  are also considered. For the continued tracking of the segmented object at Frame $t$, the state-of-the-art video object segmentation method STCN \cite{stcn} can be employed. Alternatively, this process can be split into tracking the bounding box first and then applying SAM segmentation, with the tracked box serving as the prompt. This division leads to further decisions regarding how the bounding box at Frame $t$ is detected and linked to its previous appearance, with options including the automatic 100DOH detector or the oracle-based ground truth bounding box. To connect detections, we experiment with IoUTracker  \cite{ioutracker, he2020fastreid}, StrongSORT  \cite{strongsort}, ByteTrack \cite{zhang2022bytetrack}, 
GTR~\cite{gtr} and MixFormer  \cite{mixformer}, combining both automatic and oracle-based methods in detection, segmentation, and tracking.

In contrast, HOIST-Former operates by producing a spatio-temporal binary segmentation mask for each hand-held object, treating the video as a singular entity and thus avoiding the need to distinctly separate these tasks. This sets HOIST-Former apart from the other baseline and oracle methods discussed. A notable comparison in terms of operation is the Mask2Former method, which we also benchmark against here.

\subsection{Experimental Results}

\cref{table.baselines} reports the performance of various methods for identifying, segmenting, and tracking hand-held objects. This analysis, detailed in \cref{sec:compare-methods}, includes comparisons with Mask2Former and HOIST-Former. Among the methods excluding Mask2Former and HOIST-Former, BL-A emerges as the most effective. However, BL-A is only viable when a user manually identifies and segments the hand-held object in the first frame, a task requiring significant time and effort. Choosing to only draw the bounding box leads to a notable performance drop of 7.3\% on the HOIST dataset (Method BL-B). If the initial bounding box is determined using the 100DOH detector instead of manual annotation (Method BL-C), performance drastically decreases, halving the AP. This highlights the critical need for accurate hand-held object bounding box detection, a task where the 100DOH detector falls short. The other methods in \cref{table.baselines}, following a similar approach of bounding box detection, linking, and segmentation using SAM with bounding box prompts, are also evaluated. BL-J, assuming ground truth bounding boxes, performs best among these. However, when ground truth bounding boxes are replaced by those detected by 100DOH, as in method BL-K, there is a significant decline in performance. MixFormer proved to be the most effective among various linking methods tested. 

\begin{figure*}
\centering 
    \centering
    \includegraphics[width=0.9\textwidth]{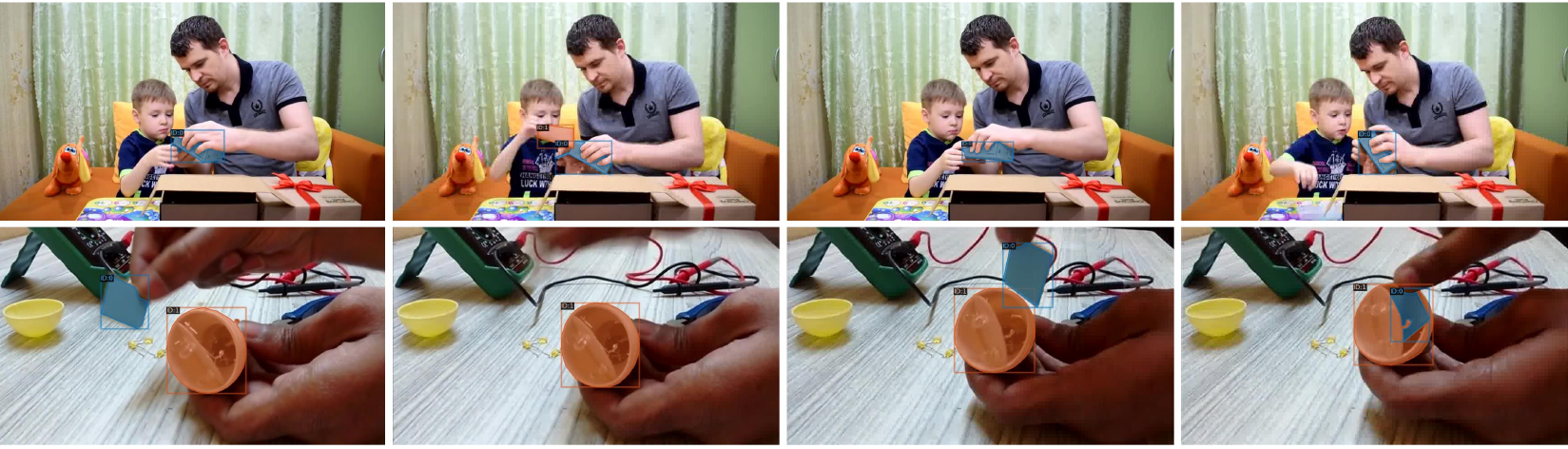}
    \vskip -0.1in
    \caption{{\bf Illustrative qualitative results of HOIST-Former}. Each row displays selected frames from a single video. Within each row, a distinct hand-held object is assigned a unique tracking ID and is consistently represented in the same color.} \label{fig.qual_res}
    \vskip -0.1in
\end{figure*}

HOIST-Former emerges as the leading method, significantly surpassing others, especially those ranging from BL-A to BL-M. This achievement is particularly noteworthy given that some methods have the advantage of accessing ground truth bounding boxes or segmentation masks, a privilege not afforded to HOIST-Former. Mask2Former ranks as the second-best method. Notably, this is not the standard Mask2Former network designed for segmenting and tracking predefined object categories. Rather, it is a re-trained version of the vanilla Mask2Former architecture, specifically adapted for a single consolidated class of hand-held objects. While Mask2Former outperforms many contenders in \cref{table.baselines}, it is still surpassed by HOIST-Former. This success of HOIST-Former is attributed to its advanced Hand-Object Transformer Decoder, which proficiently pools information and bases decisions on both hands and objects--key factors for determining hand-held status. Both Mask2Former and HOIST-Former models are trained using the HOIST training data. For the VISOR dataset, characterized by a much sparser set of video frames compared to HOIST's framerates, the models are trained on VISOR training data. The performances reported in the table for VISOR reflect this specialized training.

\vskip -0.2in

\subsection{Ablation studies and qualitative results}

A key innovation in HOIST-Former lies in its utilization of hand context to ascertain hand-held status. This novel concept is embodied in the Hand-Object Transformer Decoder, a unique Transformer decoder featuring two cross-attention modules: Hand-to-Object and Object-to-Hand, crafted for bidirectional context integration and decision-making. Additionally, the significance of this mutual context is highlighted by considering the contact boundary between hand and object segmentation, reinforced through the implementation of Contact Loss. These elements are purposefully integrated to ensure the network has sufficient information for accurate decision-making. Our ablation study, detailed in \cref{table.ablation_studies}, evaluates the criticality of these components. The results clearly demonstrate that removing either the Hand-to-Object Attention module, the Object-to-Hand Attention module, or the Contact Loss significantly impacts HOIST-Former's performance.

\begin{figure}
\centering 
    \includegraphics[width=0.9\linewidth]{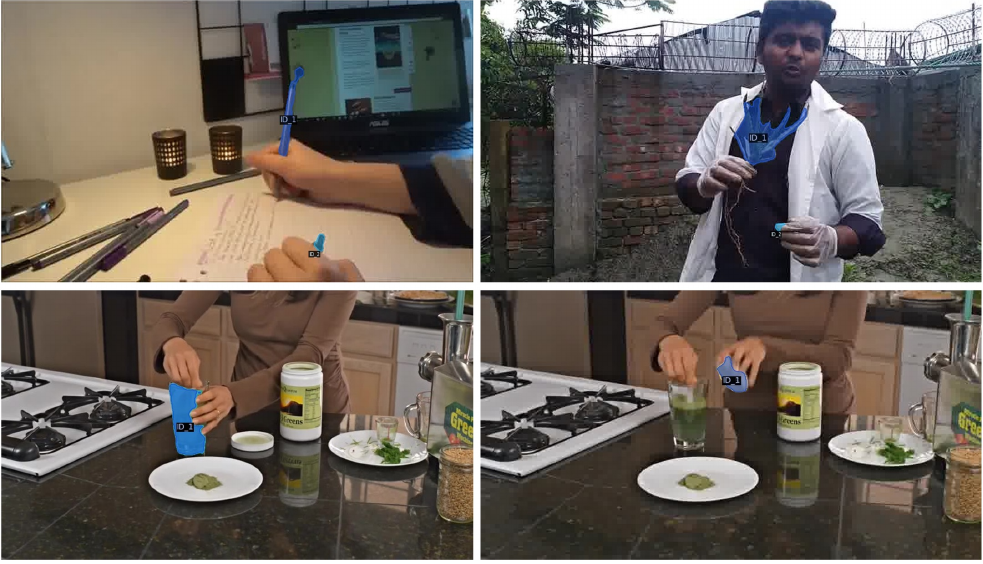}
    \vskip -0.1in

    \caption{{\bf Some failure cases of HOIST-Former}. The first row corresponds to two frames from different videos. The last row corresponds to two frames from the same video.} \label{fig.failure_cases}
\vskip -0.2in
\end{figure}

\Fref{fig.qual_res} shows results of HOIST-Former on two videos. The first row shows a case where the object is segmented only when it is hand-held. The second row shows a case where the object is asssigned the same tracking ID by HOIST-Former even after the object disappears for a while and contacts the hand later. \Fref{fig.failure_cases} shows some failure cases. The top row highlights cases where only a part of the object is segmented; the unsegmented part of the object contains extremely thin region and therefore hard to segment. The second row shows a case where a different object is assigned a previously assigned ID.

\vskip -0.25in

\section{Conclusions}

In this paper, we tackled the task of identifying, segmenting, and tracking hand-held objects. We introduced HOIST-Former, an innovative transformer-based architecture, adept at segmenting hands and objects by pooling features based on their positions and context. This approach is further refined with a contact loss that emphasizes areas where hands contact objects. We also presented the HOIST dataset, comprising 4,125 in-the-wild videos with comprehensive annotations. Our experiments on HOIST and two other public datasets showcased HOIST-Former's effectiveness in hand-held object segmentation and tracking.

{\small    \myheading{Acknowledgements.} This project was partially supported by US National Science Foundation Award NSDF DUE-2055406 and DARPA PTG HR00112220001 award. The content of the information does not necessarily reflect the position or the policy of the Government, and no official endorsement should be inferred.}

{
\small
\bibliographystyle{ieeenat_fullname}
\bibliography{main,sn_pubs}
}

\end{document}